\documentclass{article}

\usepackage{microtype}
\usepackage{graphicx}
\usepackage{subcaption}
\usepackage{booktabs}
\usepackage{hyperref}
\usepackage{xcolor} 

\usepackage[preprint]{icml2026}

\usepackage{amsmath}
\usepackage{amssymb}
\usepackage{mathtools}
\usepackage{amsthm}
\usepackage{multirow}

\usepackage[capitalize,noabbrev]{cleveref}

\theoremstyle{plain}

\theoremstyle{definition}

\theoremstyle{remark}

\icmltitlerunning{The Case for Soft-Label Training}

\begin{document}

\twocolumn[
\icmltitle{Distributions In, Distributions Out: The Case for Soft-Label Training}

\begin{icmlauthorlist}
\icmlauthor{Agamdeep Singh}{msr}
\icmlauthor{Ashish Tiwari}{msr}
\icmlauthor{Hosein Hasanbeig}{msr}
\icmlauthor{Priyanshu Gupta}{msr}
\end{icmlauthorlist}

\icmlaffiliation{msr}{Microsoft}

\icmlcorrespondingauthor{Agamdeep Singh}{t-agasingh@microsoft.com}

\icmlkeywords{Machine Learning, ICML}

\vskip 0.3in
]

\printAffiliationsAndNotice{}


\begin{abstract}
Supervised classifiers output a distribution over classes but are typically trained against a single label obtained by collapsing multiple annotators into a majority vote. On tasks where annotator disagreement reflects genuine ambiguity — natural language inference, politeness, visually ambiguous categorization — this collapse discards information and forces models to express uniform confidence on inputs where humans systematically disagree. We compare soft-label training, which uses the full annotation distribution as the target, against hard-label training across three datasets spanning vision and NLP (ChaosNLI, POPQUORN, CIFAR-10H). Soft-label training matches or exceeds hard-label accuracy on every dataset, reduces KL divergence to the annotator distribution by 32\% on average ($p < 10^{-4}$), and produces predictions whose per-sample entropy correlates 61\% more strongly with annotator entropy — models trained on distributions are uncertain precisely where humans are. We argue these benefits follow from a basic observation: when annotators legitimately disagree, the annotation distribution is the correct learning target, not a noisy estimate of it.



\end{abstract}

\begin{figure}[p]
    \centering
    \begin{subfigure}[b]{\linewidth}
        \centering
        \includegraphics[width=\linewidth]{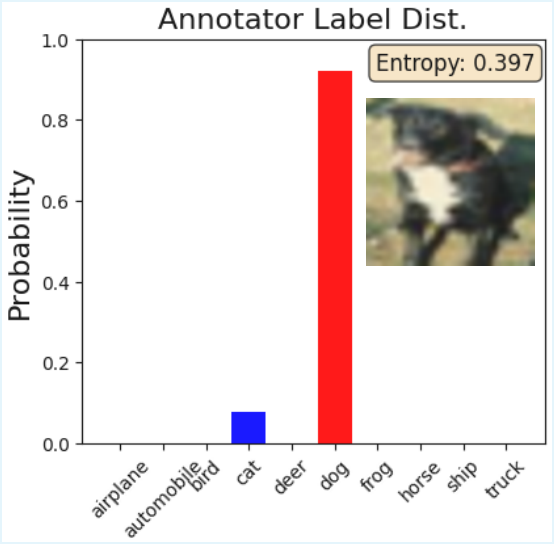}
        \caption{}
        \label{fig:low_entropy}
    \end{subfigure}
    \vspace{-0.5em}
    \begin{subfigure}[b]{\linewidth}
        \centering
        \includegraphics[width=\linewidth]{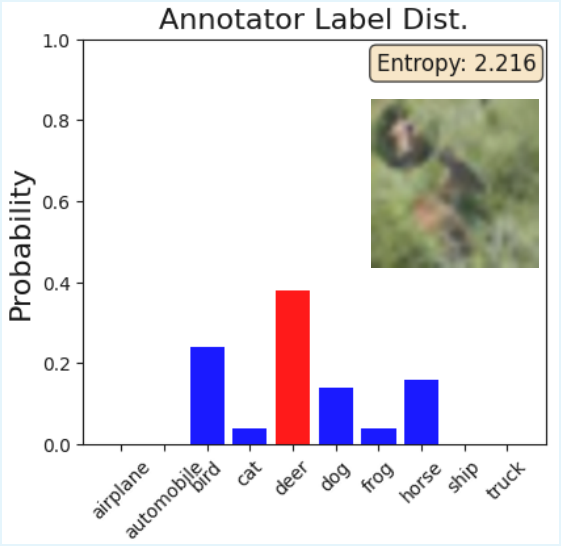}
        \caption{}
        \label{fig:high_entropy}
    \end{subfigure}
    \vspace{-0.5em}
    \caption{
        Annotator entropy varies across samples, from near-consensus (a) to substantial disagreement (b), reflecting genuinely different levels of subjective ambiguity. Images from CIFAR-10H~\cite{cifar10h}.
    }
    \label{fig:model_comparison}
\end{figure}

\section{Introduction}

Human judgment varies across many prediction tasks. What one annotator considers a polite request, another may find neutral. An ambiguous animal photo might reasonably be labeled as either a cat or dog. This variation isn't noise: it reflects genuine differences in how people perceive and categorize the world. For AI systems deployed in human contexts, alignment with this variation matters. Namely, a model that claims certainty on fundamentally ambiguous cases misrepresents reality. When annotators have subjectively varied annotations on a data sample (Figure~\ref{fig:model_comparison}), standard data pipelines collapse this distribution into a single hard label, e.g., via majority vote, before training. While neural networks inherently model predictions as probability distributions over classes ($P(Y|X=x)$), their target data is paradoxically reduced to a point estimate. We argue this disconnect flattens human diversity. In other words, when genuine ambiguity exists, the target data representation should match the probabilistic modeling. Collapsing annotations destroys information about human label variation, forcing models to express false confidence on fundamentally ambiguous samples.




\subsection{Our Position: Match the Target to the Model}

\begin{figure*}
    \centering
    \includegraphics[width=\linewidth]{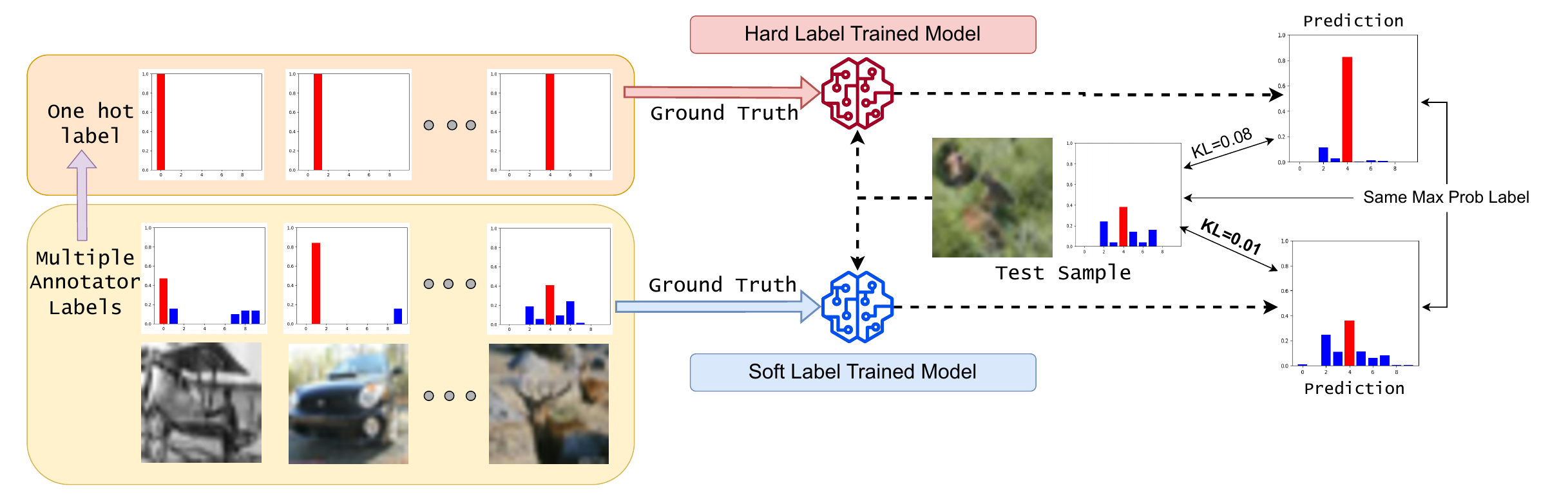}
    \caption{Both models predict the same majority class, yet their uncertainty profiles diverge sharply (KL=0.01 vs 0.08). This demonstrates that soft-label training captures the full annotation distribution rather than collapsing to artificial certainty—producing predictions that acknowledge ambiguity when it exists.}
    \label{fig:hero_diagram}
\end{figure*}

That annotators disagree is well documented. Natural language inference exhibits persistent disagreement even among expert annotators~\cite{chaosnli}, with collective opinions revealing that many examples resist consensus. Work on politeness classification~\cite{popquorn} has shown that annotator demographics meaningfully influence labeling decisions, suggesting disagreement often reflects legitimate perception differences rather than noise. Even on visual classification tasks~\cite{cifar10h}, people systematically vary in their category boundaries. Despite this recognition, the dominant paradigm treats annotation distributions as noisy measurements of a single underlying truth \cite{crowds}, applying majority voting\cite{majority_vote} or other aggregation techniques \cite{aggregation_1, aggregation_2} to extract or generate the `correct' label. We question this assumption. When expert annotators persistently disagree, their disagreement often reflects true ambiguity in the concept being labeled (e.g., Figure~\ref{fig:model_comparison}) and it is not measurement error or incompetence. Therefore, on data with genuine epistemic uncertainty, the annotation distribution \emph{is} the ground truth, not a noisy signal to be resolved. Training models to match these distributions rather than collapsed labels produces systems that are epistemically aligned with human uncertainty (Figure~\ref{fig:hero_diagram}). 


We question this practice. Modern classifiers already represent their output as a distribution $q(y \mid x)$ over classes. Yet the label they are trained against is a point, i.e., a single class index obtained by collapsing the annotation distribution. The mismatch is in how the data is \emph{represented} for training, not in what the model can express. When annotators systematically disagree (e.g., Figure~\ref{fig:model_comparison}), the empirical annotation distribution is a far better estimate of $p(y \mid x)$ than any single annotator's vote. Training $q(y \mid x)$ against the empirical distribution rather than its argmax aligns the target representation with the model's representation, and produces systems that track human disagreement (Figure~\ref{fig:hero_diagram}). This alignment has practical benefits for model robustness, calibration, and trustworthy deployment.


Soft-label training is well established as an effective technique for distribution alignment. It has been widely used in knowledge distillation~\cite{hinton2015distillingknowledgeneuralnetwork} and label distribution learning~\cite{geng2016labeldistributionlearning}. Recent work investigates bias and noise in soft labels~\cite{NEURIPS2023_bad8ddae,ML2025}, methods for eliciting them~\cite{DBLP:conf/emnlp/WuLMSBS23}, and theoretical advantages for ambiguous data~\cite{pmlr25}. However, these works frame soft labels as a training technique that happens to work well.

Our contribution reframes the question: we argue that when data exhibits genuine uncertainty, annotation distributions are not approximations, they \emph{are} the correct target. This perspective explains why soft-label training's known benefits (improved calibration, reduced overfitting, better generalization) emerge: they are natural consequences of training on correct targets rather than collapsed approximations.

\subsection{Findings}

We find that soft-label training (using full annotation distributions) compared to hard-label training (using collapsed majority votes) yields:
\begin{itemize}
    \item \textbf{Better distributional alignment:} Lower KL divergence between model predictions and human annotation distributions
    \item \textbf{Improved robustness:} Soft-label models train longer without overfitting
    \item \textbf{Maintained accuracy:} No degradation in standard accuracy metrics
    \item \textbf{Tracks human disagreement:} Per-sample model entropy correlates more strongly with annotator entropy

\end{itemize}

We validate these findings across three datasets spanning vision and NLP domains, showing that the benefits hold consistently when distributions are treated as ground truth.

\section{Measuring Distributional Alignment}
To train and evaluate against annotation distributions, we need metrics that operate on distributions rather than collapsed labels, aside from accuracy. We use two complementary measures.

\paragraph{KL divergence.} For a test sample with annotation distribution $p$ and model prediction distribution $q$ over $C$ classes, we report the Kullback--Leibler divergence ~\cite{kld}.
\begin{equation}
    D_{KL}(p \,\|\, q) = \sum_{c=1}^{C} p_c \log \frac{p_c}{q_c}.
    \label{eq:kl}
\end{equation}

We use forward KL, $D_{KL}(p \,\|\, q)$, rather than reverse KL, $D_{KL}(q \,\|\, p)$, because the two have qualitatively different behavior on multi-modal targets~\cite{minka2005divergence}. Forward KL is \emph{mode-covering}: the penalty at each class $c$ is weighted by $p_c$, so any class with nonzero annotator mass that the model assigns near-zero probability incurs arbitrarily large loss. Its optimum is $q = p$, requiring the model to reproduce the full annotation distribution. Reverse KL, by contrast, is \emph{mode-seeking}: its penalty is weighted by $q_c$, so the model is free to place zero mass on classes that annotators selected, and its optimum on a non-degenerate $p$ collapses $q$ onto a single mode. For a distribution such as $p = (0.6, 0.3, 0.1)$, forward KL is minimized at $q = p$, while reverse KL is minimized at $q = (1, 0, 0)$---precisely the hard-label collapse we argue against. Forward KL is therefore the only choice consistent with our position, and it coincides (up to the constant $H(p)$) with the soft-label cross-entropy used during training:
\begin{equation}
    H(p, q) = -\sum_{c=1}^{C} p_c \log q_c = H(p) + D_{KL}(p \,\|\, q),
    \label{eq:ce_kl}
\end{equation}
ensuring training objective and evaluation metric are aligned.
A lower KLD indicates better distributional alignment---the model's predictive distribution is closer to the annotator distribution. We report mean KLD across the test set alongside standard accuracy.

\paragraph{Entropy correlation.} KLD captures distributional distance, but does not directly tell us whether the model is \emph{more} uncertain on samples where annotators disagree more. To measure this, we compute the Pearson correlation between the entropy of the annotator distribution and the entropy of the model predicted distribution across the test set. For a distribution $r$, entropy is
\begin{equation}
    H(r) = -\sum_{c=1}^{C} r_c \log r_c,
    \label{eq:entropy}
\end{equation}
and the per-sample correlation across $N$ test samples is
\begin{equation}
    \rho = \frac{\sum_{i=1}^N \left(H(p^{(i)}) - \bar{H}_p\right)\left(H(q^{(i)}) - \bar{H}_q\right)}{\sqrt{\sum_{i=1}^N \left(H(p^{(i)}) - \bar{H}_p\right)^2 \, \sum_{i=1}^N \left(H(q^{(i)}) - \bar{H}_q\right)^2}},
    \label{eq:entropy_corr}
\end{equation}
where $\bar{H}_p$ and $\bar{H}_q$ denote the test-set means of $H(p^{(i)})$ and $H(q^{(i)})$ respectively. A higher $\rho$ indicates that model uncertainty tracks human disagreement on a per-sample basis.

\section{Experimental Setup}

\subsection{Datasets}

\begin{figure}[]
    \centering
    \includegraphics[width=\columnwidth]{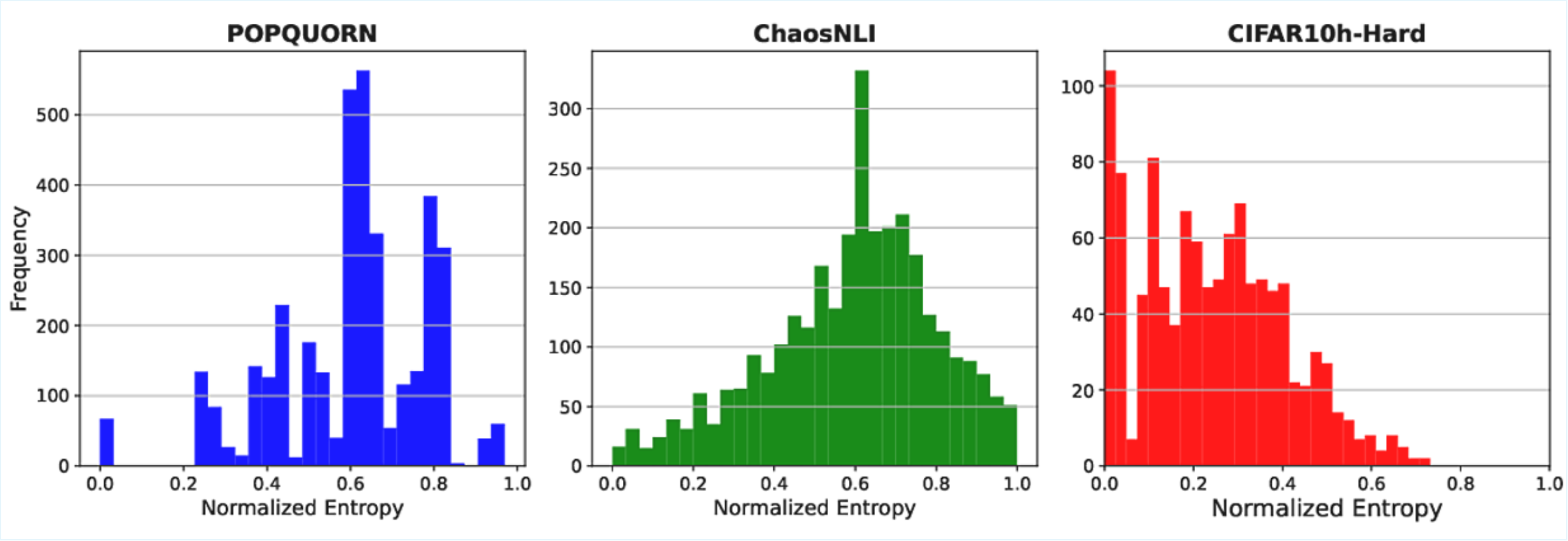}
    \caption{Entropy distributions across datasets. Entropy is normalized for number of classes.}
    \label{fig:entropy_dist}
\end{figure}

We perform an evaluation using three datasets with substantial annotation disagreement, spanning both the NLP and vision domains. Table~\ref{tab:datasets} presents some statistics of the data sets. Figure~\ref{fig:entropy_dist} shows the entropy distributions for all three datasets, demonstrating substantial variation in annotation uncertainty within each dataset. The three datasets are:

\begin{table}[h]
\centering
\caption{Dataset characteristics. All datasets exhibit high annotation spread. Cls=classes, Ann/Smp = Annotators per Sample.}
\label{tab:datasets}
\small 
\setlength{\tabcolsep}{4pt} 
\begin{tabular}{@{}lcrcc@{}}
\toprule
\textbf{Dataset} & \textbf{Cls} & \textbf{N} & \textbf{Ann/Smp} & \textbf{Mean Entropy} \\ \midrule
ChaosNLI       & 3  & 3,113 & 100   & $0.59$  \\
POPQUORN       & 5  & 3,700 & 6.7   & $0.62$ \\
CIFAR-10H-H    & 10 & 1,103 & 50    & $0.24$ \\ \bottomrule
\end{tabular}
\end{table}

\textbf{ChaosNLI \cite{chaosnli}:} A natural language inference dataset with soft labels representing multiple human annotators' judgments on premise-hypothesis pairs classified into three categories: entailment, neutral, or contradiction.

\textbf{POPQUORN \cite{popquorn}:} A politeness classification dataset with labels representing varying degrees of politeness in text.

\textbf{CIFAR-10H-Hard:} A curated subset of CIFAR-10H \cite{cifar10h}, which models human categorization over a large behavioral dataset, comprising more than 500,000 judgments over 10,000 natural images from ten object categories. We used stratified uniform sampling with 10 entropy bins and up to 200 samples per bin (or less if unavailable), resulting in 1,103 samples.

Figure~\ref{fig:entropy_dist} shows the entropy distributions for all three datasets, demonstrating substantial variation in annotation uncertainty within each dataset.

\subsection{Implementation Details}

For all experiments, we used frozen pretrained embeddings with lightweight MLP classification heads to isolate the effect of label type (soft vs. hard) from representation learning. For text classification tasks, we employed OpenAI Text Embeddings 3 Large as the frozen embedding model, while for image classification, we used DINOv2 Small. In both cases, the classification head consisted of a 2-layer MLP with dropout and GELU activation. Only the classification head parameters were trained, keeping the pretrained embeddings fixed throughout all experiments.
\subsection{Training Procedure}

For soft-label training, we use a cross-entropy loss where the full human annotation distribution serves as the target distribution. For hard-label training, we collapse this distribution to its majority-vote class and apply the standard categorical cross-entropy loss. All other training procedures, model architectures, and optimization settings are identical across both conditions to ensure a fair comparison.

\paragraph{Hyperparameter Search and Training Setup.}
To enable robust and unbiased evaluation, we conducted an exhaustive grid search over key hyperparameters. The search space included learning rate $\in \{10^{-3}, 10^{-4}, 10^{-5}\}$, batch size $\in \{8, 16, 32\}$, number of epochs $\in \{10, 15, 20\}$, weight decay $\in \{0, 10^{-4}\}$, and scheduler type $\in \{\text{None}, \text{ReduceLROnPlateau}\}$. Each configuration was trained independently for both soft- and hard-label variants using identical train/validation/test splits.

Early stopping was applied with a patience of 5 epochs and a minimum validation loss improvement threshold of $\Delta = 10^{-4}$. When active, the \texttt{ReduceLROnPlateau} scheduler reduced the learning rate by a factor of 0.5 upon stagnation in validation loss. All experiments were implemented in PyTorch using the Adam optimizer. Model selection was performed based on the minimum validation loss, and all best configurations were subsequently retrained with 10 random seeds to assess statistical significance. During training, we monitored loss, accuracy, and Kullback–Leibler (KL) divergence across the training, validation, and test sets to evaluate both predictive performance and distributional alignment. Table~\ref{tab:hyperparameters} shows the best-performing hyperparameters selected for each dataset and model type.


\begin{table}[]
\centering
\caption{Best-performing hyperparameters selected via grid search.}
\label{tab:hyperparameters}
\small
\setlength{\tabcolsep}{5pt} 
\begin{tabular}{llccccc}
\toprule
\textbf{Dataset} & \textbf{Mode} & \textbf{LR} & \textbf{BS} & \textbf{Ep.} & \textbf{WD} & \textbf{Sched.} \\ \midrule
ChaosNLI    & Soft & $10^{-3}$ & 32 & 20 & $10^{-4}$ & Plat. \\
            & Hard & $10^{-4}$ & 16 & 15 & $10^{-4}$ & Plat. \\ \midrule
POPQUORN    & Soft & $10^{-5}$ & 8  & 20 & 0         & None  \\
            & Hard & $10^{-4}$ & 32 & 10 & 0         & Plat. \\ \midrule
CIFAR-10H-H & Soft & $10^{-4}$ & 16 & 15 & $10^{-4}$ & Plat. \\
            & Hard & $10^{-3}$ & 16 & 10 & $10^{-4}$ & None  \\ \bottomrule
\end{tabular}
\end{table}

\section{Results}

\subsection{Training Dynamics and Overfitting}

\begin{figure*}
    \centering
    \includegraphics[width=\linewidth]{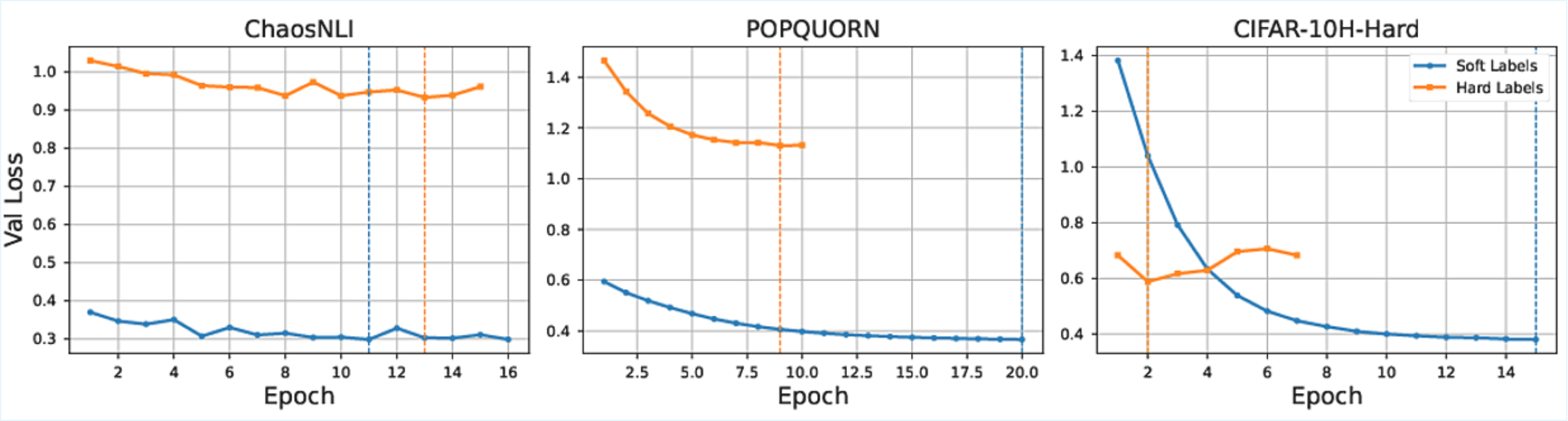}
    \caption{Validation loss for soft-label (blue) and hard-label (orange) models. Loss magnitudes differ due to different target representations (distributions vs. one-hot) but training dynamics are revealing: soft-label models maintain stable or improving validation performance longer, while hard-label models plateau or degrade earlier.}
    \label{fig:training_curves}
\end{figure*}

Figure~\ref{fig:training_curves} illustrates the validation loss trajectories for both soft-label and hard-label models across all three datasets. For hard-label training, validation loss begins increasing earlier in training, indicating overfitting to the collapsed labels. Soft-label training, by contrast, can train for more epochs while validation performance continues to improve. This suggests that distributional targets provide a more informative learning signal that regularizes the model.

\subsection{Distributional Alignment and Accuracy}

Table~\ref{tab:soft_vs_hard_results} summarizes the final test performance across all datasets. We report mean and standard deviation across 10 random seeds, with paired t-tests to assess statistical significance.

\begin{table*}[]
\centering
\caption{
Comparison of soft-label and hard-label training across datasets.
Values show mean ± std over 10 seeds.
$t$-tests assess significance between model pairs ($\alpha=0.05$).
Bold indicates the better value (higher accuracy, lower KL). 
}
\scriptsize
\label{tab:soft_vs_hard_results}
\resizebox{\textwidth}{!}{
\begin{tabular}{lcccccc}
\toprule
\textbf{Dataset} & \textbf{Model} & \textbf{Accuracy (\%)} & \textbf{KL} & \textbf{$t$-test (Acc) $p$} & \textbf{$t$-test (KL) $p$} & \textbf{Significant}\\
\midrule
\multirow{2}{*}{\texttt{cifar10h\_hard}} 
& Hard & 78.51 ± 2.47 & 0.596 ± 0.073 & \multirow{2}{*}{0.553} & \multirow{2}{*}{1.6e-05} & \multirow{2}{*}{KL}\\
& Soft & 78.14 ± 2.52 & \textbf{0.406 ± 0.021} & & & \\ 
\midrule
\multirow{2}{*}{\texttt{chaos\_nli}} 
& Hard & 51.75 ± 1.25 & 0.367 ± 0.011 & \multirow{2}{*}{9.1e-04} & \multirow{2}{*}{4.7e-07} & \multirow{2}{*}{Acc, KL}\\
& Soft & \textbf{55.30 ± 2.05} & \textbf{0.319 ± 0.012} & & & \\ 
\midrule
\multirow{2}{*}{\texttt{popquorn}} 
& Hard & 52.00 ± 1.25 & 0.430 ± 0.014 & \multirow{2}{*}{0.948} & \multirow{2}{*}{6.4e-09} & \multirow{2}{*}{KL }\\
& Soft & 51.98 ± 1.46 & \textbf{0.361 ± 0.007} & & & \\ 
\bottomrule
\end{tabular}}
\end{table*}

Three key findings emerge:

\textbf{(1) Distributional alignment improves significantly:} Soft-label training achieves significantly lower KL divergence on all three datasets ($p < 0.001$), with an average improvement of 32\%. This indicates that soft-label models better capture human uncertainty.

\textbf{(2) Accuracy is maintained or improved:} Soft-label training does not sacrifice predictive accuracy. On ChaosNLI, it significantly improves accuracy (55.30\% vs. 51.75\%, $p < 0.001$). On CIFAR-10H-Hard and POPQUORN, accuracies are statistically indistinguishable.

\textbf{(3) Benefits are consistent across domains:} The advantages of soft-label training hold for both NLP (ChaosNLI, POPQUORN) and vision (CIFAR-10H-Hard) tasks. Figure \ref{fig:dist_Example}(Appendix) illustrates this effect with an example: soft-label predictions mirror the spread of human annotations, while hard-label predictions collapse to single peaks despite genuine annotator disagreement.

\subsection{Correlation with Data Uncertainty}


\begin{table}[h]
\centering
\caption{Pearson correlation between human annotation entropy and model prediction entropy.}
\label{tab:entropy_correlation}
\small
\begin{tabular}{lccc}
\toprule
\textbf{Dataset} & \textbf{Hard} & \textbf{Soft} & \textbf{$\Delta$} \\ \midrule
ChaosNLI       & 0.130 & \textbf{0.284} & +119\% \\
CIFAR-10H-Hard & 0.353 & \textbf{0.493} & +40\%  \\
POPQUORN       & 0.232 & \textbf{0.286} & +24\%  \\ \bottomrule
\end{tabular}
\end{table}

Table~\ref{tab:entropy_correlation} presents the correlation between model prediction entropy and annotation distribution entropy. On average, predictions from soft-label trained models show 61\% stronger entropy correlation with the data. This effect is most pronounced on ChaosNLI (+119\%), where high annotation rates (100 per sample) provide the most reliable distributional targets. Even on POPQUORN, with fewer annotations per sample, soft-label training improves correlation by 24\%.

This demonstrates that soft-label training not only reduces overall distributional distance (KL divergence) but also produces models whose uncertainty appropriately tracks the inherent ambiguity of each example.

\section{Discussion}


Our findings have practical consequences for developing reliable AI systems:

\textbf{Evaluation:} Judging models solely on hard-label accuracy penalizes honest uncertainty expression on ambiguous data. A model that confidently predicts the majority label on a 60-40 split may achieve higher accuracy but worse human distribution alignment. Namely, a model that says "60\% confident" on ambiguous cases is more valuable than one that falsely claims 100\% certainty. 




\textbf{Data collection} The most common objection to soft-label training is that multi-annotator data is rarely available---and this is true today precisely because the field has standardized on single-label collection and training pipelines. We argue this is a self-reinforcing convention rather than a fundamental constraint. Our results show meaningful gains at modest annotation rates (6.7 per sample on POPQUORN), with the effect scaling further when more annotations are available (100 per sample on ChaosNLI). For tasks involving subjective or contested judgments---safety, toxicity, politeness, preference, NLI---we recommend treating multi-annotator collection as the default. We acknowledge that multi-annotation data is currently scarce and typically more expensive to collect than standard single-label datasets. 





The main limitation of our work is the limited number of benchmark sets used for evaluation. However, the diversity of the datasets and the strength of the observed signal provides us confidence that the observations also hold for other datasets. We also acknowledge that our results are for classification tasks. The idea can be extended to other learning tasks and generative models (eg. LLMs), and warrants more exploration in those domains.

\section{Conclusion}

Standard supervised classification trains and evaluates models that output a distribution over classes against targets that are a single class index, often obtained by collapsing multi-annotator data into a single label. We have argued that on tasks where annotators legitimately disagree, the annotation distribution itself is the correct learning target, and that collapsing it to a point estimate discards information about genuine ambiguity in the data. Across three datasets spanning vision and NLP, soft-label training matches or exceeds hard-label accuracy, reduces KL divergence to the annotator distribution by 32\% on average, and produces models whose per-sample predictive entropy correlates substantially more strongly with annotator entropy. The benefits are consistent across domains and hold even at modest annotation rates. Soft-label training is not new as a technique, but our results suggest its advantages on subjectively ambiguous data are not incidental regularization effects: they follow from training on a correctly represented target. For tasks involving contested or subjective human judgments, we recommend that multi-annotator collection and distributional training become the default.

For trustworthy AI, we need systems that know what they don't know, and what is inherently subjective. Training on distributional ground truth is a crucial step toward that goal.

\bibliography{references}
\bibliographystyle{icml2026}
\appendix
\section{Appendix}
\begin{figure*}[]
    \centering
    \includegraphics[width=0.9\linewidth]{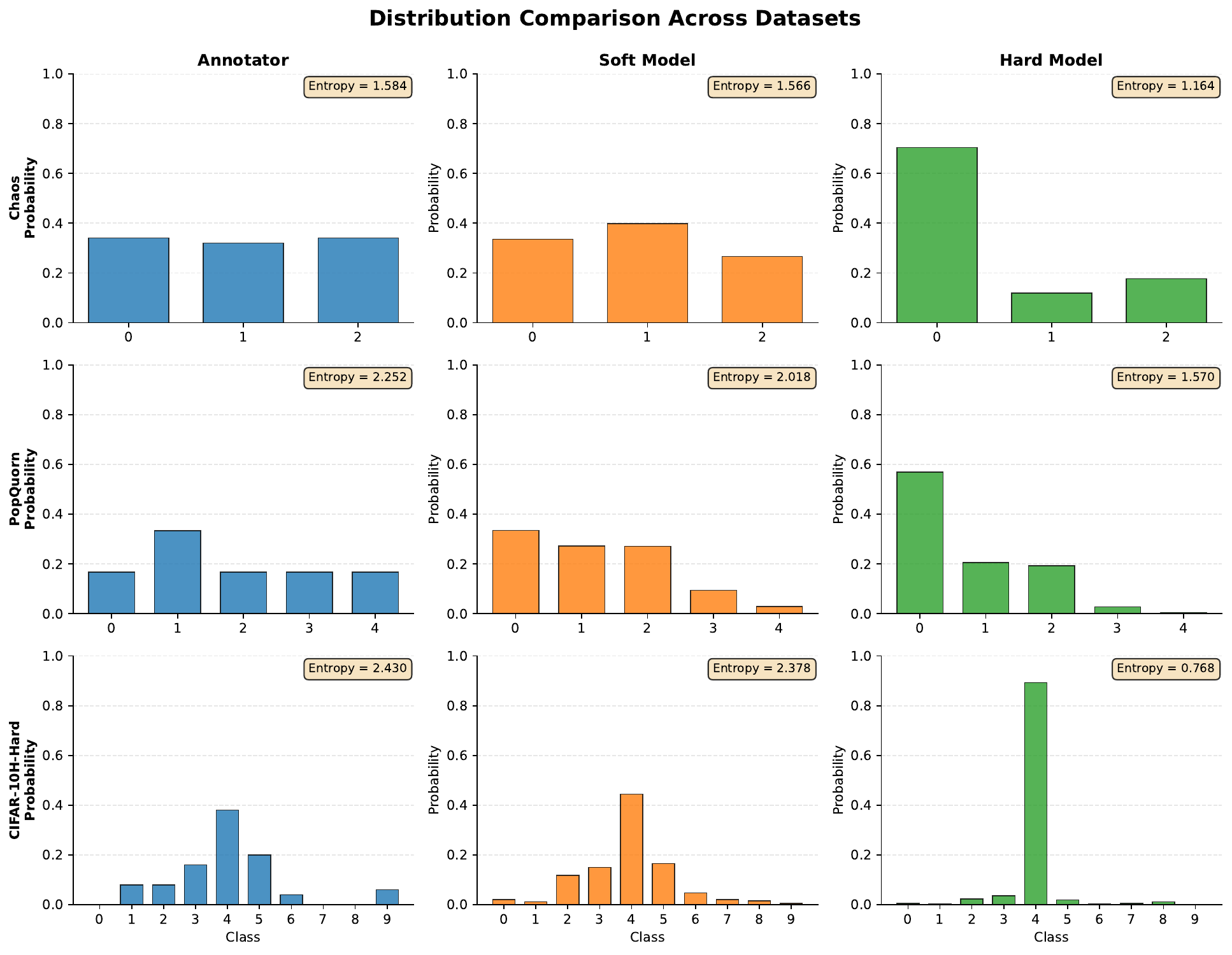}
    \caption{On high-entropy samples, soft-label predictions (orange) mirror the spread of human annotations (blue), while hard-label predictions (green) collapse to single peaks despite genuine disagreement. This shows that training method directly shapes whether models express or suppress uncertainty—affecting trustworthiness on fundamentally ambiguous inputs.}
    \label{fig:dist_Example}
\end{figure*}

Figure~\ref{fig:dist_Example} picks one example each from the three datasets and shows the label distribution (in blue on the left) for that example, followed by the distribution predicted by the model trained using soft-label training (in orange in the middle), and the distribution predicted by the model trained using hard-label training (in green on the right) for that example.
\end{document}